\documentclass[journal]{IEEEtran}

\usepackage{url}
\usepackage{times}
\usepackage{epsfig}
\usepackage{amssymb}
\usepackage{subfigure}
\usepackage{bbding}
\usepackage{xcolor}
\usepackage{threeparttable}
\usepackage{booktabs}
\usepackage{algorithm}
\usepackage{algorithmic}

\newcommand{\eg}{\textit{e}.\textit{g}.}
\newcommand{\ie}{\textit{i}.\textit{e}.}
\newcommand{\et}{\textit{e}\textit{t} \textit{a}\textit{l}.}

\usepackage{graphicx, color, multirow, amsmath}

\RequirePackage{latexsym,amsmath,amssymb}

\ifCLASSINFOpdf
\else
\fi

\begin{document}

\title{Viewport-Unaware Blind Omnidirectional Image Quality Assessment: A Unified and Generalized Approach}
\author{
Jiebin Yan,
Kangcheng Wu,
Jingwen Hou,
Jiayu Zhang,
Pengfei Chen,
Yuming Fang,~\IEEEmembership{Senior~Member,~IEEE}

\thanks{This work was supported in part by the National Natural Science Foundation of China under Grants U24A20220, 62461028, 62132006, 62301378, and 62501257, in part by the Natural Science Foundation of Jiangxi Province of China under Grants 20243BCE51139, 20252BAC230003, 20223AEI91002, and 20252BAC200010, and and in part by the Early-Career Young Scientists and Technologists Project of Jiangxi Province under grant 20252BEJ730134. (Corresponding author: Jingwen Hou).}

\thanks{Jiebin Yan and Jingwen Hou are with the School of Computing and Artificial Intelligence, Jiangxi University of Finance and Economics, Nanchang 330032, Jiangxi, China, and also with the Jiangxi AI Quality Testing and Inspection Center, Nanchang 330029, Jiangxi, China. (e-mail: yanjiebin@jxufe.edu.cn, jingwen003@e.ntu.edu.sg).}

\thanks{Kangcheng Wu, Jiayu Zhang and Yuming Fang are with the School of Computing and Artificial Intelligence, Jiangxi University of Finance and Economics, Nanchang 330032, Jiangxi, China. (e-mail: kangchengwu-my@foxmail.com, jiayuzhang7@foxmail.com, fa0001ng@e.ntu.edu.sg).}

\thanks{Pengfei Chen is with the School of Artificial Intelligence, Xidian University, Xi'an, China (e-mail: chenpengfei@xidian.edu.cn).}

}


\maketitle

\begin{abstract}
Blind omnidirectional image quality assessment (BOIQA) presents a great challenge to the visual quality assessment community, due to different storage formats and diverse user viewing behaviors. The main paradigm of BOIQA models includes two steps, \ie, viewport generation, and quality prediction, which brings an extra computational burden and is hard to generalize to other visual contents (\eg, 2D planar image). Thus, in this paper, we make an attempt to solve these issues. First, we experimentally find that BOIQA can be formulated as a blind (2D planar) image quality assessment (BIQA) problem, \ie, the first step - viewport generation - is no longer needed, which narrows the natural gap between BOIQA and BIQA. Then, we present a new BOIQA approach, which has three merits: \ie, \emph{viewport-unaware} - it accepts an omnidirectional image in the widely used equirectangular projection format as input without any transformation; \emph{unified} - it can also be applied to BIQA; and \emph{generalized} - it shows better generalizability against other competitors. Finally, we validate its promise by held-out test, cross-database validation, and the well-established gMAD competition. The source code is available at \url{https://github.com/KangchengWu/VUGA}.
\end{abstract}


\section{Introduction}
\label{sec:intr}

\IEEEPARstart{I}{n} the era of continuous innovation in visual imaging, the omnidirectional image (OI) has become indispensable in various fields, \eg, virtual reality (VR), augmented reality (AR), autonomous driving, remote monitoring, \emph{etc.} However, OIs are prone to quality degradation during generation, compression, transmission, and storage~\cite{wang2017begin,fang2017no,fang2017objective,yan2020no}, which can lead to issues such as noise, resolution loss, compression artifacts, \emph{etc.}, all of which may have a negative effect on quality of experience (QoE)~\cite{fang2020cvpr}. Therefore, accurate omnidirectional image quality assessment (OIQA) has become crucial to ensure optimal QoE and facilitate downstream applications~\cite{yan2023survey}. The OIQA studies can be categorized into subjective study and objective method, where the former often refers to building subjective databases through psychophysical experiments to explore how those influencing factors interact with subjective ratings, and the latter represents the design of mathematical models to quantify the quality of OIs. Depending on the availability of reference images, the OIQA methods can be roughly classified into full-reference OIQA (FR-OIQA) and no-reference/blind OIQA (BOIQA). Compared to FR-OIQA methods~\cite{yan2025towards}, the BOIQA methods can be deployed without access to reference images, which makes them more practical. 

\begin{figure}[t]
\centering
\includegraphics[scale=0.32]{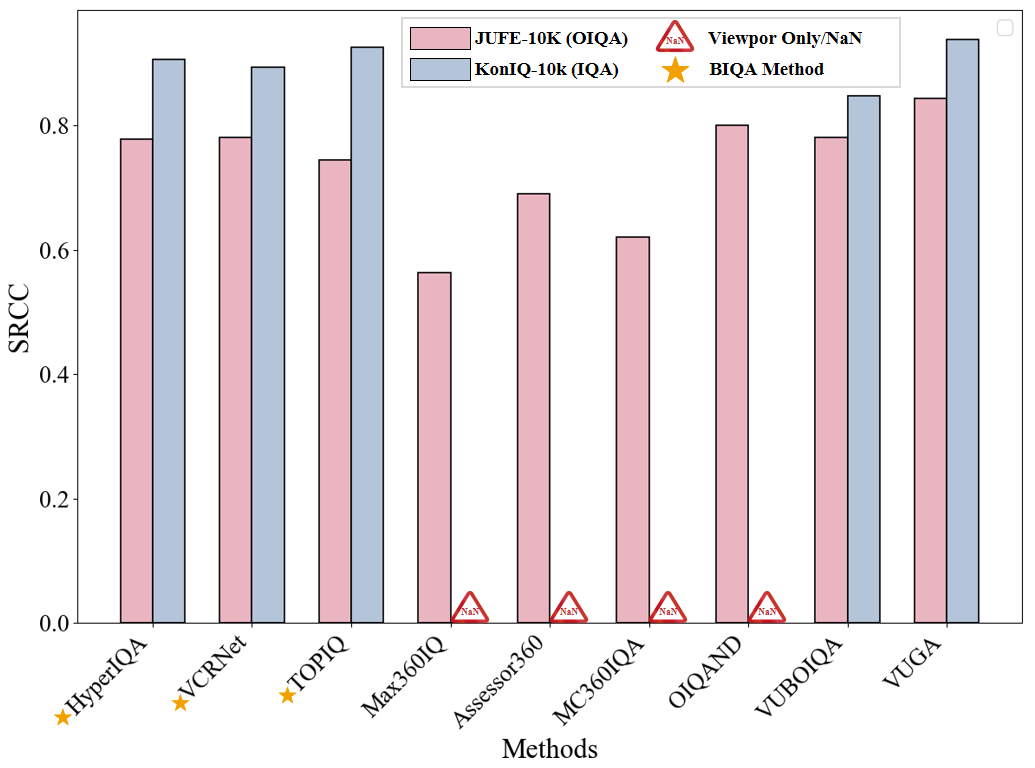}
\caption{The performance comparison of BOIQA models (MC360IQA~\cite{sun2019mc360iqa}, Assessor360~\cite{wu2024assessor360}, OIQAND~\cite{yan2024subjective}, VUBOIQA~\cite{yan2025vuboiqa} and the proposed VUGA) and BIQA models (VCRNet~\cite{pan2022vcrnet}, TOPIQ~\cite{chen2024topiq} and HyperIQA~\cite{su2020blindly}) on the JUFE-10K and KonIQ-10k databases, where these BIQA models and the proposed VUGA accept raw ERP image as input, VUBOIQA uses the patch sequence from ERP image as input, and the rest models accept the viewport sequence as input. ``NaN'' denotes that the performance is incalculable due to the fundamental architectural gap between BOIQA and BIQA.}
\label{fig:gap}
\end{figure} 

\IEEEpeerreviewmaketitle

Early OIQA methods~\cite{chen2018spherical,zhou2018weighted,zakharchenko2016quality,sun2017weighted} mainly rely on 2D planar image quality assessment (IQA) methods (``2D'' is omitted for simplicity), such as peak signal-to-noise ratio (PSNR) and structural similarity (SSIM)~\cite{wang2004image}, by considering the spherical geometry characteristic of OIs. These methods advance OIQA to some extent; however, their performance is suboptimal due to the limited representation ability. To address the aforementioned issue, many deep learning-based OIQA methods have been proposed~\cite{liu2024perceptual}. One of the main paradigms is the so-called \textit{viewport-aware} model, which is based on the fact that a user can only see a limited field of view (\ie, viewport or patch) at a certain moment, and such a model extracts a viewport sequence (also can be regarded as a pseudo 2D video) from an OI as input~\cite{sui2021perceptual,sui2023scandmm}. Since the formats of OIs are varied, including equirectangular projection (ERP), pyramid projection (PYM)~\cite{li2023mfan}, cubemap projection (CMP)~\cite{jiang2021cubemap}, pseudo-cylindrical projection~\cite{cai2024pseudocylindrical}, \emph{etc.}, those objective BOIQA models usually extract viewports from a test OI in one format (\eg, ERP) or a combination of these formats, combining the generated viewports with a deep learning model for end-to-end quality prediction~\cite{yan2025vuboiqa}.

Although significant progress has been made in BOIQA, several challenges remain: 
\romannumeral1) \textbf{Most existing methods rely on viewport generation}, following a two-step paradigm (viewport generator and quality assessor). This introduces computational overhead, as viewport processing adds additional computational burden to fixed-architecture models. Yan~\et~\cite{yan2025vuboiqa} attempted to address this by extracting a patch sequence directly from the raw ERP image, eliminating viewport generation. However, this method~\cite{yan2025vuboiqa} still requires repeated feature calculations for different viewports. \romannumeral2) \textbf{A fundamental architectural gap exists between BOIQA and BIQA}, preventing direct transfer of the success of BIQA~\cite{Chen_2025_CVPR} to BOIQA. As shown in Fig.~\ref{fig:gap}, BIQA models can be seamlessly adapted to OIQA without modifying themselves, while BOIQA models require architectural modifications (except VUBOIQA). In order to tackle these two challenges, we propose a Viewport-unaware, Unified, and Generalized BOIQA Approach (VUGA).  Specifically, VUGA employs 1) a backbone with frozen parameters for hierarchical feature extraction, 2) a cross-scale multi-receptive-field perception (CMP) module (plugged between backbone layers) that distills quality-aware features from the inborn geometric structures via parallel architecture, 3) an adaptive feature fusion (AFF) module for progressively merging hierarchical quality-sensitive features, 4) a quality regression module for mapping those merged features to quality score. 

In summary, our key contributions are threefold.

\begin{itemize}

\item We present the first empirical study on narrowing the natural gap between BOIQA and BIQA. Our findings reveal that BOIQA can be formulated as a BIQA problem; specifically, the previous viewport-dependent paradigm can be transferred to the mature 2D solution.

\item We propose a novel viewport-unaware BOIQA method that progressively distills hierarchical quality-aware features from inborn geometric structures. To achieve this, we design an effective parallel model using deformable convolutions and stacked convolutions with different receptive fields to capture distortion from both local and global perspectives.

\item We conduct extensive experiments on eight OIQA datasets and five IQA databases. The results show that the proposed method is applicable to BOIQA and BIQA tasks, while achieving superior performance against the state-of-the-art OIQA and BIQA methods.
\end{itemize}

\section{Related Work}
\label{sec:rw}

\subsection{BIQA Models}
BIQA aims to predict image quality without a reference image, and traditional BIQA methods~\cite{fang2017no,yan2020no} rely on hand-crafted features. Although they effectively capture synthetic distortion and are computationally efficient, they struggle with complex real-world distortion. Deep learning-based IQA methods broke through the limitations of traditional approaches by automatically learning image features end-to-end. These methods typically involve some pre-processing of the original image, such as patch sampling, cropping, and resizing. Kang~\et~\cite{kang2014convolutional} early introduced deep learning in the BIQA task, where image patches are extracted after local contrast normalization and image quality is predicted end-to-end. Su~\et~\cite{su2020blindly} proposed a self-adaptive hyper network, which takes random sampling and horizontal flipping of the input image and predicts image quality by combining local and global features. Pan~\et~\cite{pan2022vcrnet} proposed a BIQA method based on visual restoration, where the original image is first resized and cropped, then image quality is predicted using a visual restoration network and a quality estimation network. Song~\et~\cite{unicontent} proposed a unified BIQA method, in which the original image is scaled and cut into patches. Chen~\et~\cite{chen2024topiq} proposed a top-down BIQA method, where the image is resized directly and used as input, and high-level semantic information is used to guide the extraction of low-level distortion features.

These patch-based strategies (\eg, patch sampling and cropping) can be regarded as data augmentation methods and are suitable for images with uniform distortion. However, they cannot be used for non-uniformly distorted images (\ie, different regions may suffer from varying amounts of distortion~\cite{yan2020no}), making it unreasonable to assign a global score to different patches. Therefore, existing BOIQA models usually process the extracted viewports or patches together, leading to a fundamental architecture gap between BOIQA and BIQA.

\begin{figure*}[t]
\centering
\includegraphics[scale=0.56]{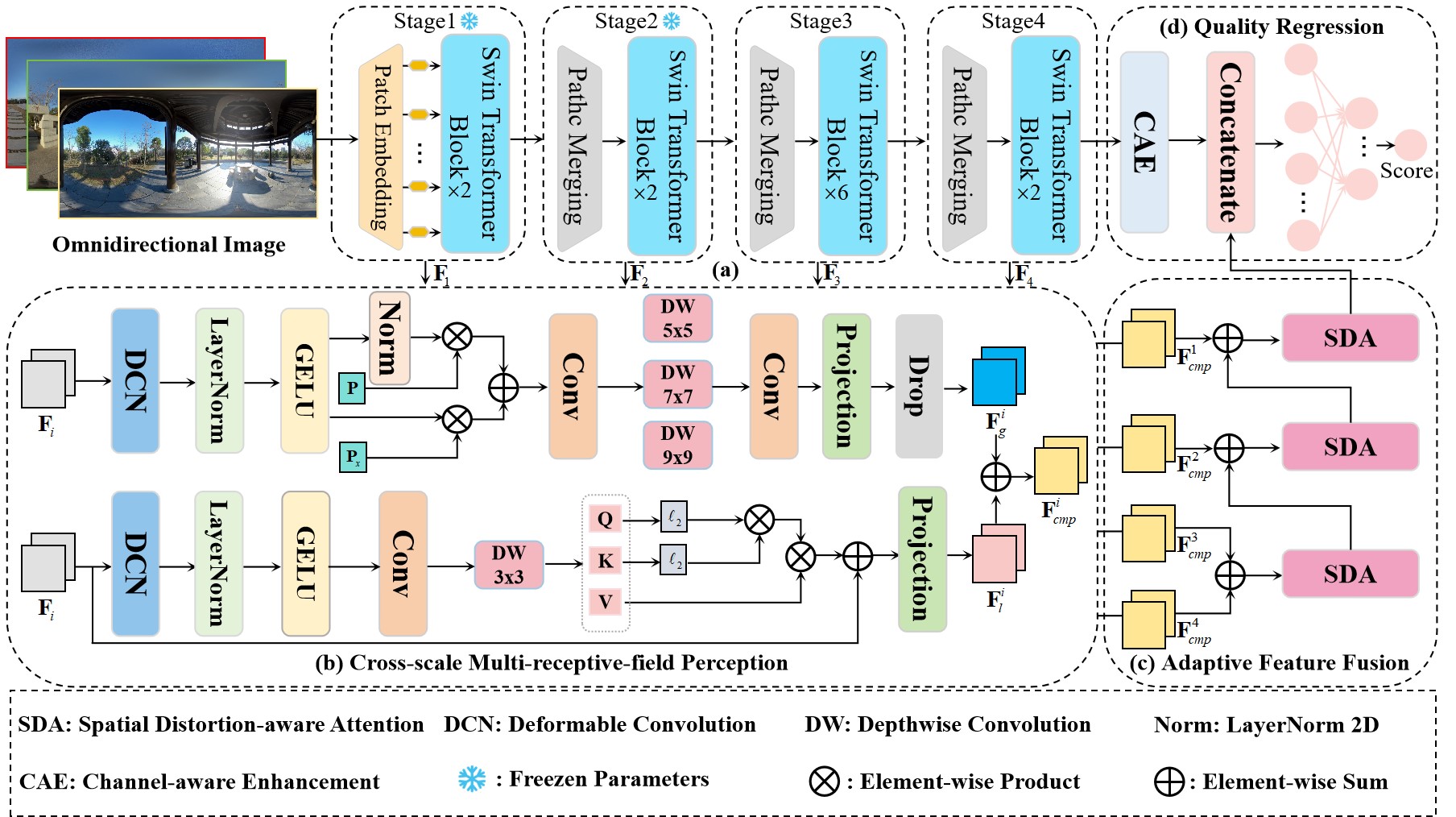}
\caption{The framework of the proposed VUGA. It consists of four parts: (a) backbone with frozen parameters for extracting hierarchical features, (b) cross-scale multi-receptive-field perception module for distilling quality-aware features from the inborn geometric structures, (c) adaptive feature fusion module for progressively merging those hierarchical quality-sensitive features, and (d) quality regression module for mapping those merged features to quality score.}
\label{fig:framework}
\end{figure*}

\subsection{OIQA Models}

A straightforward way to design OIQA models is to extend conventional IQA metrics to accommodate the spherical projection characteristic of OIs. Representative approaches include PSNR-based methods such as S-PSNR~\cite{yu2015framework}, WS-PSNR~\cite{sun2017weighted}, and CPP-PSNR~\cite{zakharchenko2016quality}, as well as SSIM-based methods such as S-SSIM~\cite{chen2018spherical} and WS-SSIM~\cite{zhou2018weighted}. With the rise of deep learning, numerous OIQA methods have emerged. Based on how they process OIs, these methods can be categorized into viewport-aware and viewport-unaware methods. Viewport-aware models primarily perform viewport generation based on viewing priors or scan-path prediction. Sun~\et~\cite{sun2019mc360iqa} sampled six viewports from each OI and employed a multichannel convolutional neural network to predict quality by averaging their features. Xu~\et~\cite{xu2020vgcn} proposed a dual-branch BOIQA model that fuses local and global features to predict quality. Fang~\et~\cite{fang2022perceptual} proposed a BOIQA method with multiscale feature extraction and quality prediction, incorporating user viewing behaviors and conditions into viewport features. Wu~\et~\cite{wu2024assessor360} proposed a recursive sampling method to simulate user browsing through pseudo-viewport sequences, and used multiscale aggregation and temporal modeling to fuse viewport features. Yan~\et~\cite{yan2024subjective} simulated user browsing by extracting equatorial viewports and weighting them based on the degree of non-uniform distortion. Yan~\et~\cite{yan2025matoiqa} introduced three quality-related auxiliary tasks with task-specific feature fusion to assess non-uniformly distorted OIs.

Unlike viewport-aware models, viewport-unaware models always convert OIs into different projection formats as input. Jiang~\et~\cite{jiang2021cubemap} proposed a perception-driven OIQA framework based on CMP. The framework combines head and eye movement-based attention features with six cubemap features of OI, and predicts quality scores using three special schemes. Li~\et~\cite{li2023mfan} proposed multi-projection fusion attention OIQA method, which reduces projection distortions by combining three projection types (including CMP, PYM, and ERP) for quality prediction. Yan~\et~\cite{yan2025vuboiqa} sampled several image patches from the ERP image using equatorial prior knowledge and calculated quality scores with patch-level attention.

In summary, viewport-aware BOIQA methods perform well, but require additional computations for viewport prediction (if exists) and repeated processing of viewports. In contrast, existing viewport-unaware BOIQA methods simplify the pipeline but fail to address geometric deformation in ERP images, resulting in suboptimal performance.

\section{The proposed method}
In this section, we first outline the problem formulation. Subsequently, the proposed method VUGA, as shown in Fig.~\ref{fig:framework}, is thoroughly described according to its each module.

\subsection{Problem Formulation}

Given a raw ERP image $\mathbf{I}$ (which can be replaced by a 2D planar image when adapted to BIQA), our objective is to learn a model $\mathcal{M}(\cdot)$ : $\mathbb{R}^{C\times H \times W} \rightarrow \mathbb{R}$, where $C$ represents the number of channels, $H$ is the height and $W$ is the width of $\mathbf{I}$. The
above process can be described as:
\begin{equation}
    \hat{s} =  \mathcal{M}(\mathcal{R}(\mathbf{I},\textit{o}), \boldsymbol{\theta}),
\label{eq:formulation}
\end{equation}
where $\mathcal{R}$ is the preprocessing operator and $o$ is the scaling factor, $\boldsymbol{\theta}$ represents the learnable parameters, $\hat{s}$ is the predicted quality score.

In fact, equation~\ref{eq:formulation} is suitable for BOIQA and BIQA, making it possible to design a unified solution. However, OIs show a completely different geometry distribution with 2D planar images, therefore we elaborately design several specific modules to constitute $\mathcal{M}$. We then introduce these modules one by one.

\begin{figure}[]
\centering
\includegraphics[scale=0.26]{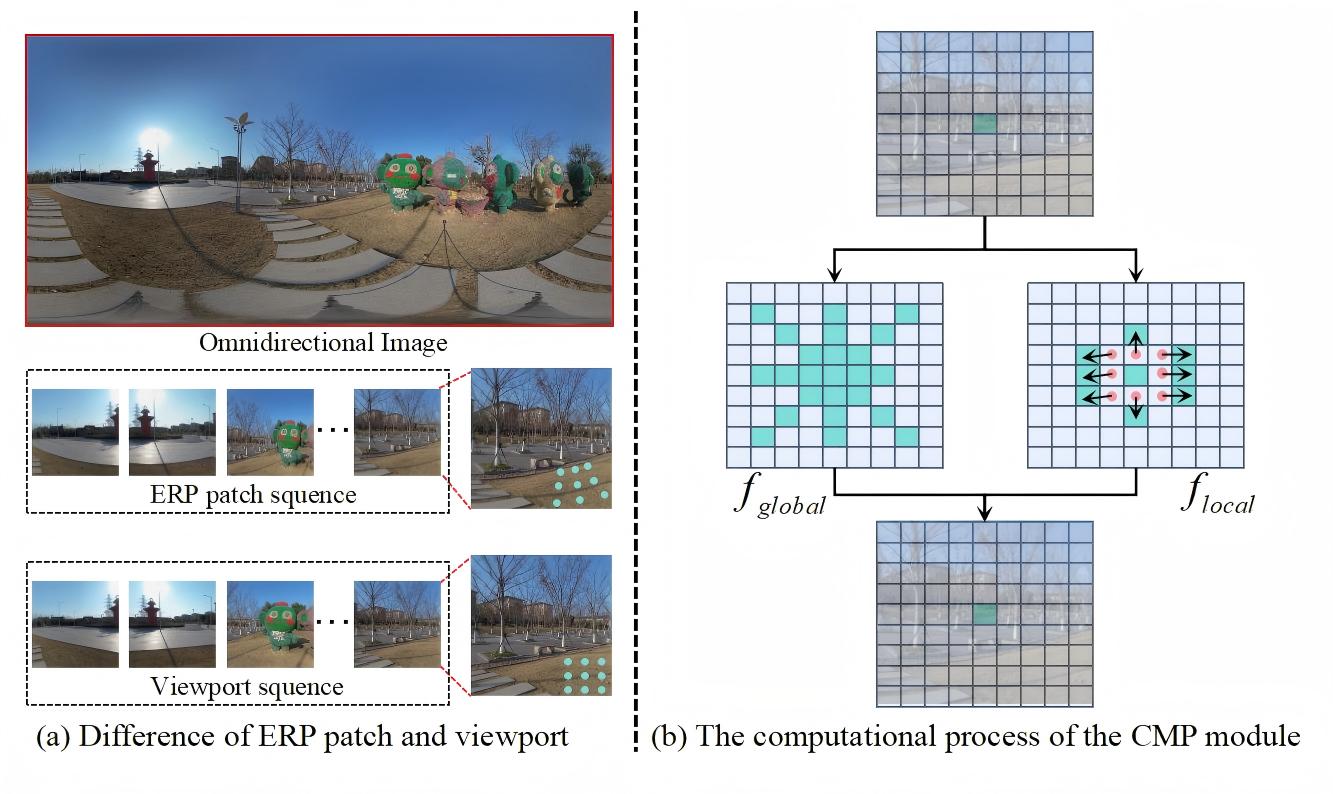}
\caption{The geometry deformation in omnidirectional images and the computational process of CMP module.}
\label{fig:geometry}
\end{figure} 

\subsection{Hierarchical Feature Extraction}

The Swin Transformer V2 Tiny (SwinV2-T)~\cite{liu2022swin} is adopted as the backbone for hierarchical feature extraction. Its shifted window-based self-attention effectively models spatial dependencies and distortion patterns, making it well suited for both 2D planar images and OIs. For the preprocessed image $\mathbf{\widehat{I}}$, we extract multiscale feature representations from the four stages of the backbone capturing both low-level perceptual representations and high-level semantic structures, which can be formulated as follows:
\begin{equation}
 \mathbf{F}_i = \{f_{b}(\mathbf{\widehat{I}};\boldsymbol{\theta}_{b})\}_{i=1}^4,
\end{equation}
where $\mathbf{F}_i$ represents the feature map from the $i$-th stage of the backbone, $f_{b}$ and $\boldsymbol{\theta}_{b}$ denote the backbone and its parameters, respectively.

\subsection{Cross-scale Multi-receptive-field Perception}
Since ERP images suffer from inherent irregular geometric deformation induced by projection, this fact poses significant challenges to unified modeling with conventional methods. To address this issue, we propose a Cross-scale Multi-receptive-field Perception (CMP) module, which integrates deformable convolution (DCN)~\cite{zhu2019deformable} with a hybrid local-global attention mechanism, enabling efficient distillation of distortion from geometry deformation, as shown in Fig.~\ref{fig:geometry}. Specifically, each hierarchical feature map $\mathbf{F}_i$ is refined using DCN, which adaptively adjusts sampling positions, thereby alleviating projection-induced irregular geometric deformation, which can be described as:
\begin{equation}
\mathbf{D}_i = \sigma_g(\mathcal{N}(\mathcal{D}(\mathbf{F}_i)),
\end{equation}
where $\mathbf{D}_i \in \mathbb{R}^{C \times H \times W}$ denotes the feature representation refined through distortion-aware adaptation, $\sigma_g$ denotes the Gaussian error linear unit (GELU), $\mathcal{N}(\cdot)$ denotes the layer normalization operation, and $\mathcal{D}$ denotes the DCN with $3 \times 3$ kernel size. In the local branch, $\mathbf{D}_{i}$ is further processed to more effectively capture non-uniform distortion information from local regions. Specifically, we apply a $1\times1$ convolution to project $\mathbf{D}_{i}$ into a feature map of dimensionality $\mathbb{R}^{3C \times H \times W}$. Then a $3\times3$ depthwise convolution (DWConv)~\cite{howard2017mobilenets} with a dilation rate of 2 is applied to expand the local receptive field and enhance the model’s sensitivity to distorted regions. The resultant feature map is evenly split along the channel dimension into three separate components, corresponding to $\mathbf{Q}_i$, $\mathbf{K}_i$, and $\mathbf{V}_i$, each with a dimensionality of $\mathbb{R}^{C \times H \times W}$. For better stability in the computation of attention weights, we apply the $\ell_2$ normalization to both $\mathbf{Q}_i$ and $\mathbf{K}_i$ along the channel dimension. Finally, the features obtained from the local interaction modeling are fused with the original input feature through a residual connection, and then mapped by a $1 \times 1$ convolution to form the final local distortion-aware feature $\mathbf{F}_l^i \in \mathbb{R}^{512 \times H \times W}$, which can be written as:
\begin{equation}
\mathbf{F}_l^i = \mathcal{C}_{1 \times 1} \bigg( 
\frac{ \exp( \frac{ \mathbf{Q}_i \cdot \mathbf{K}_i^{T} }{ \sqrt{d} } ) }
{ \sum_j \exp ( \frac{ \mathbf{Q}_i \cdot \mathbf{K}_i^{T}[j] }{ \sqrt{d} }) } 
\cdot \mathbf{V}_i 
\bigg),
\end{equation}
where $\mathcal{C}_{1 \times 1}$ denotes the $1 \times 1$ convolution, $\text{exp}(\cdot)$ denotes the exponential function, $\sum_j$ denotes a summation over all key positions for the current query, $d$ denotes the dimension of $\mathbf{Q}_i$ or $\mathbf{K}_i$. In the global branch, we normalize the input feature $\mathbf{D}_{i}$, and use the learnable parameters $\mathbf{P}$ and $\mathbf{P}_x$ to adaptively fuse the normalized feature with the original feature. This design further handles irregular geometric deformation in the polar regions to obtain the feature map $\mathbf{\Phi}_i  \in \mathbb{R}^{C \times H \times W }$. The process can be described as: 
\begin{equation}
\mathbf{\Phi}_i =  (\lambda \cdot \frac{\textbf{D}_i - \mu}{\sqrt{\upsilon^2 + \varepsilon}} + \beta) \cdot \mathbf{P} + \textbf{D}_i \cdot \mathbf{P}_x,
\end{equation}
where $\mu$ and $\upsilon$ denote the mean and variance computed along the spatial dimensions $(H, W)$ of each feature channel, $\varepsilon$ is a small constant, $\lambda$ and $\beta$ denote learnable parameters. To reduce computational burden, $\mathbf{\Phi}_i$ is reshaped to $C \times (H \times W)$. Then, we reduce the number of channels to one-fourth of the original using a linear projection and adjust its dimensions to obtain the feature vector $\mathbf{R}_i \in \mathbb{R}^{C//4 \times H \times W}$. Subsequently, we input $\mathbf{R}_i$ into the multiscale convolution module. This module employs three parallel depthwise convolutions with receptive fields of $5\times5$, $7\times7$, and $9\times9$, enhancing the model's ability to comprehensively perceive large-scale uniform distortion and semantic information in global regions. The three parallel convolution outputs are averaged along the channel dimension. The whole process can be described as:
\begin{equation}
\mathbf{Y}_i = \frac{1}{3} \sum_{k \in \{5, 7, 9\}} \mathcal{W}_{k\times k}(\mathbf{R_i}), 
\end{equation}
where $\mathbf{{Y}}_i$ denotes the feature map from the multiscale convolution module, $\sum$ denotes the summation function, $\mathcal{W}_{k \times k}$ denotes the DWConv with $k\times k$ kernel size. Followed by a $1\times1$ convolution to adjust channels of $\mathbf{{Y}}_i$, which is then connected to the input feature $\mathbf{D}_i$ via a residual connection, and its channels are projected into 512 dimensions to obtain the global distortion-aware feature $\mathbf{F}_g^i \in \mathbb{R}^{512 \times H \times W}$. Finally, an element-wise summation operation is applied between $\mathbf{F}_g^i$ and $\mathbf{F}_l^i$ to ensure that the model can simultaneously focus on both global uniform distortion and local non-uniform distortion features, yielding the output of the CMP module, $\mathbf{F}_{cmp}^i$.

\subsection{Adaptive Feature Fusion}
\begin{figure}[t]
\centering
\includegraphics[scale=0.22]{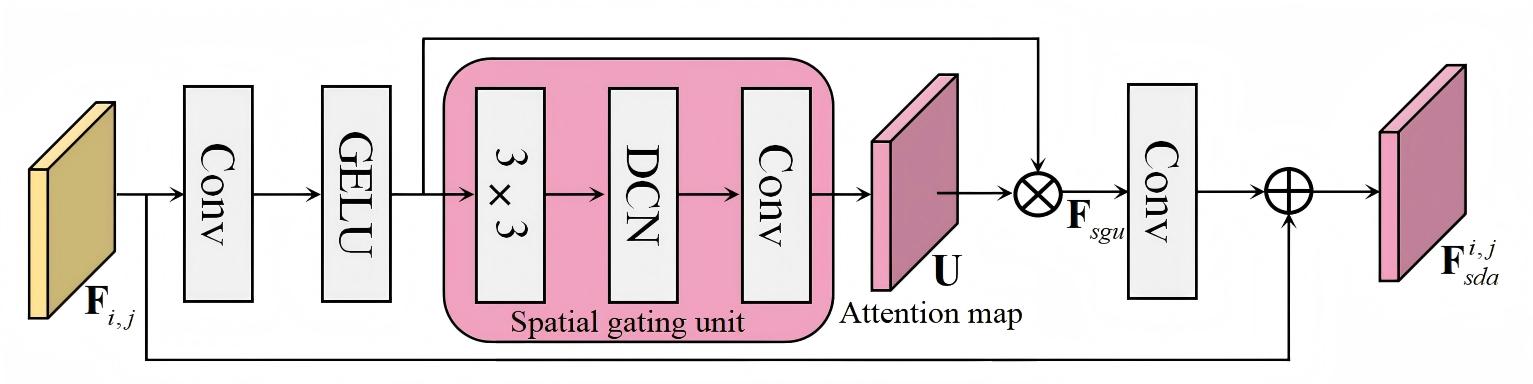}
\caption{The architecture of spatial distortion-aware attention module. The 3$\times$3 denotes a $3\times3$ depthwise convolution.}
\label{fig:SDA}
\end{figure}

Furthermore, we design an Adaptive Feature Fusion (AFF) module to integrate hierarchical distortion-aware features $[\mathbf{F}_{cmp}^1,\mathbf{F}_{cmp}^2,\mathbf{F}_{cmp}^3,\mathbf{F}_{cmp}^4]$ generated by feeding $\mathbf{D}_i$ into the CMP module. Firstly, we adopt a top-down progressive feature fusion strategy, starting with the integration of $\mathbf{F}_{cmp}^3$ and $\mathbf{F}_{cmp}^4$ to obtain $\mathbf{F}_{4,3}$. However, due to the limited spatial detail in deep features, direct addition may suppress distortion-relevant local semantic information. To address this, we introduce a Spatial Distortion-aware Attention (SDA) module,  as shown in Fig.~\ref{fig:SDA}. By combining DWConv with a DCN-based gating mechanism, this module adaptively highlights distortion-relevant regions, thereby improving the model’s capability to perceive complex distortion. In particular, given an input feature $\mathbf{F}_{4,3}$, it is first projected into an intermediate feature representation $\mathbf{F}_{4,3}'$, which is then fed into a spatial gating unit to produce the dynamic spatial attention map $\mathbf{U}$. The process can be formulated as:
\begin{equation}
\mathbf{U} = \mathcal{D}(\mathcal{W}(\mathbf{F}_{4,3}').
\end{equation}
The output is then projected through a $1\times1$ convolution and element-wise multiplied with $\mathbf{F}_{4,3}'$ to generate the output of the spatial gating unit $\mathbf{F}_{sgu}$. To preserve the original semantic information, a residual connection is introduced by adding $\mathbf{F}_{4,3}$ to $\mathbf{F}_{sgu}$. This operation can be formulated as:
\begin{equation}
    \mathbf{F}_{sda}^{4,3} = \mathcal{C}_{1\times1}(\mathbf{F}_{\text{sgu}}) + \mathbf{F}_{4,3},
\end{equation}
where $\mathbf{F}_{sda}^{4,3}$ denotes the final output feature obtained by processing the $\mathbf{F}_{4,3}$ through the SDA module. After obtaining the high-level feature $\mathbf{F}_{sda}^{4,3}$, we continue the top-down fusion process by progressively integrating shallower features. At each stage, the deeper feature is first up-sampled via bilinear interpolation, then element-wise added to the corresponding shallower feature, and subsequently refined by a SDA module to enhance distortion-aware representations. This process can be formulated as:
\begin{equation}
\begin{cases}
\mathbf{F}_{sda}^{2,3,4} &= \text{SDA}\left(\mathbf{F}_{cmp}^2 + \text{Up}(\mathbf{F}_{sda}^{4,3})\right), \\
\mathbf{F}_{aff} &= \text{SDA}\left(\mathbf{F}_{cmp}^1 + \text{Up}(\mathbf{F}_{sda}^{2,3,4})\right),
\end{cases}
\end{equation}
where $\mathbf{F}_{sda}^{2,3,4}$ denotes the resultant feature representation by fusing the feature $\mathbf{F}_{cmp}^2$ with the up-sampled $\mathbf{F}_{sda}^{4,3}$, followed by enhancement through the SDA module, $\text{Up}$ denotes an upsampling operation performed via a bilinear interpolation with a scale factor of 2, $\mathbf{F}_{aff}$ denotes the output of the AFF module.

\begin{figure}[t]
\centering
\includegraphics[scale=0.19]{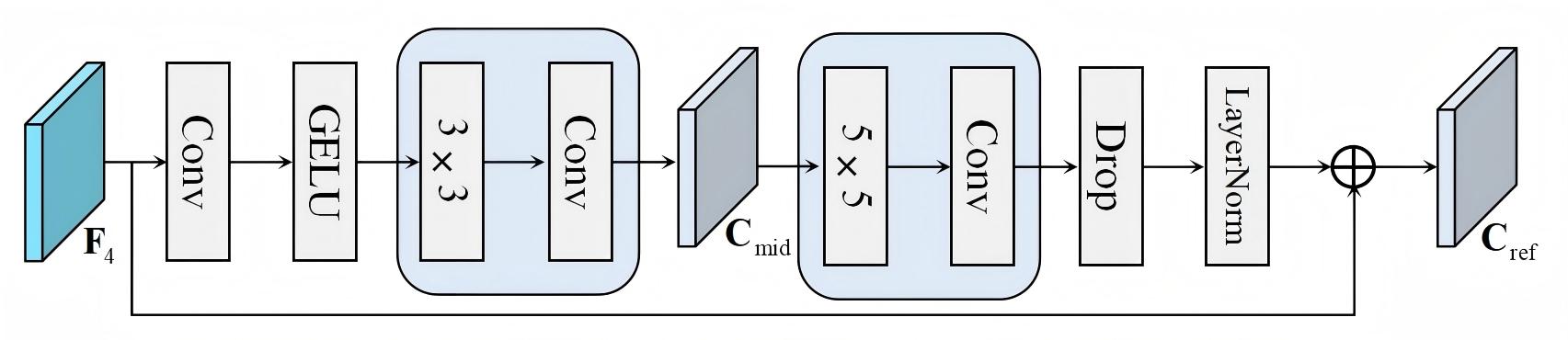}
\caption{The architecture of channel-aware enhancement module. The 5$\times$5 denotes a $5\times5$ depthwise convolution.}
\label{fig:cae}
\end{figure} 

\begin{table*}[h]
\begin{threeparttable}
\caption{Performance of general networks and BIQA models on OIQA databases. The best results are highlighted in bold.}
\renewcommand{\arraystretch}{1.5}
\setlength{\tabcolsep}{1.43mm}
\begin{tabular}{ccccccccccccccccc}
\hline
\multirow{2}{*}{Method} & \multicolumn{2}{c}{CVIQ} & \multicolumn{2}{c}{OIQA} & \multicolumn{2}{c}{MVAQD} & \multicolumn{2}{c}{IQA-ODI} & \multicolumn{2}{c}{OSIQA} & \multicolumn{2}{c}{AIGCOIQA} & \multicolumn{2}{c}{JUFE-10K} & \multicolumn{2}{c}{OIQ-10K} \\ \cline{2-17} 
                        & SRCC        & PLCC       & SRCC        & PLCC       & SRCC        & PLCC        & SRCC         & PLCC         & SRCC        & PLCC        & SRCC          & PLCC         & SRCC          & PLCC         & SRCC         & PLCC         \\ \cline{1-17}
ResNet-50~\cite{he2016deep}& 0.944       & 0.969      & 0.874       & 0.879      & 0.913       & 0.935       & 0.913        & 0.935        & 0.903       & 0.906       &     0.734          &       0.718       & 0.792         & 0.791        & 0.719        & 0.720        \\
ResNet-152~\cite{he2016deep}               & 0.949       & 0.974      & 0.895       & 0.891      & 0.809       & 0.805       & 0.890        & 0.917        & 0.908       & 0.897       & 0.837        &    0.752          & 0.793         & 0.790        & 0.731        & 0.734        \\
SwinV2-T~\cite{liu2022swin}                & 0.955       & 0.978      & 0.923       & 0.919      & 0.914       & 0.921       & 0.942        & 0.968        & 0.920       & 0.920       & 0.764              & 0.654             & \textbf{0.806}         & \textbf{0.806}        & 0.765        & 0.772        \\
SwinV2-B~\cite{liu2022swin}               & 0.946       & 0.976      & 0.937       & 0.938      & 0.904       & 0.914       & 0.938        & 0.967        & 0.922       & \textbf{0.934}       &  0.658             &  0.770           & 0.797         & 0.799        & 0.795        & 0.802       \\  \cline{1-17}
HyperIQA~\cite{su2020blindly}                & 0.976       & 0.959      & 0.890       & 0.893      & 0.846       & 0.836       & 0.902        & 0.889        & 0.919       & 0.914       &  0.815             & 0.734             & 0.778         & 0.778        & 0.738        & 0.738        \\ 
VCRNet~\cite{pan2022vcrnet}                & 0.947       & 0.965      & 0.943       & 0.942      & 0.846       & 0.872       & 0.848        & 0.923        & \textbf{0.939}       & 0.914       &  0.755             & 0.651             & 0.781         & 0.777        & 0.736        & 0.734        \\
TOPIQ~\cite{chen2024topiq}                &0.618  & 0.701   & 0.854    &0.893  &0.719  &0.780 &0.123  &0.229  &0.857  &0.862  & 0.464   &0.394 &0.745  &0.744   &0.695   & 0.704 \\ 
QualiClip~\cite{agnolucci2024quality}                &0.188  & 0.307   & 0.330    &0.329  &0.246  &0.249 &0.154  &0.147  &0.749  &0.819  & 0.727   &0.628 &0.310  &0.314   &0.132   & 0.133 \\ \cline{1-17}
HyperIQA*~\cite{su2020blindly}       &0.960   &  0.973  & \textbf{0.968}  & \textbf{0.970}   &0.817  &0.822  & 0.887  & 0.929  & 0.820 & 0.717  & \textbf{0.936}  &   \textbf{0.923}   &  0.763  & 0.763  & \textbf{0.802}  & \textbf{0.805}             \\ 
VCRNet*~\cite{pan2022vcrnet}          & 0.866   & 0.830  & 0.884  & 0.880   &  0.862  &   0.843   &0.819   &  0.738   &0.356     &0.307   &0.512    & 0.477     & 0.712  & 0.737   &0.751   &0.803              \\
TOPIQ*~\cite{chen2024topiq}                &0.975  &0.980    &0.931     &0.902  & 0.838 &0.850 &0. 600  &0.585  &  0.645 &0.783  &0.548    &0.542 &0.555  &0.587  &0.770  & 0.782 \\ 
QualiClip*~\cite{agnolucci2024quality}                 &\textbf{0.978}    &\textbf{0.982}  &0.878  & 0.883  & 0.843  & 0.850  &\textbf{0.981}   & \textbf{0.987}   & 0.604  &0.572    & 0.897   &0.895    &0.211   & 0.318    & 0.395  &0.414              \\ 
\bottomrule
\end{tabular}

\begin{tablenotes}
\item[] The * denotes that these methods use a viewport sequence (eight viewports) as input.
\end{tablenotes}

\label{cross domain}
\end{threeparttable}
\end{table*}

\subsection{Quality Regression}
To enhance the model's ability to perceive and represent distortion-related features, we introduce the Channel-Aware Enhancement (CAE) module, as shown in Fig.~\ref{fig:cae}, to refine the high-level feature map \(\mathbf{F}_4\). Although $ \mathbf{F}_4 $ contains rich semantic information, it exhibits limited sensitivity to distortion-relevant features. The CAE module applies DWConv with varying receptive fields and channel-wise projections to adaptively enhance distortion-aware features while preserving semantic integrity. The whole process can be described as:
\begin{equation}
\begin{cases}
\mathbf{C}_{\text{mid}} &= \mathcal{C}_{1 \times 1}\left(\mathcal{W}_{3 \times 3}\left(\sigma_g\left(\mathcal{C}_{1 \times 1}(\mathbf{F}_4)\right)\right)\right), \\
\mathbf{C}_{\text{ref}} &= \text{Dropout}\left(\mathcal{C}_{1 \times 1}\left(\mathcal{W}_{5 \times 5}\left(\mathbf{C}_{\text{mid}}\right)\right)\right),
\end{cases}
\end{equation}
where $\mathbf{C}_{\text{mid}}$ denotes the intermediate feature representation obtained through the first-stage enhancement, $\mathcal{W}_{3 \times 3}$ and $\mathcal{W}_{5 \times 5}$ denote DWConv with $3\times3$ and $5 \times 5$ kernels, respectively; $\mathbf{C}_{\textit{ref}}$ denotes the refined feature map. After obtaining $\mathbf{C}_{\textit{ref}}$, we apply layer normalization and a residual connection to obtain the final result $\mathbf{F}_{cae}$. To construct a unified quality-aware representation, the flattened features $\mathbf{F}_{cae}$ and $\mathbf{F}_{aff}$ are concatenated:
\begin{equation}
    \mathbf{V}_{\text{quality}}= \text{Concat}(\text{Flatten}(\mathbf{F}_{cae}),\text{Flatten}(\mathbf{F}_{aff})),
\end{equation}
where $\mathbf{V}_{\text{quality}}$ denotes the concatenated feature vector used for quality regression, $\text{Concat}(\cdot)$ denotes the concatenation operation; $\text{Flatten}(\cdot)$ denotes the flattening operation. The vector $\mathbf{V}$ is passed through a two-layer fully connected (FC) regressor with ReLU activation to predict the final perceptual quality score, which can be described as:

\begin{equation}
    \hat{s}_n =  \text{FC}(\mathbf{V}_{\text{quality}}, \boldsymbol{\theta}_{f}),
\end{equation}
where $\hat{s}$ denotes the predicted quality score of the $n$-th ERP image, $\boldsymbol{\theta}_{f}$ denotes the learnable parameters of the FC layer. In this study, the model is trained and optimized using the mean squared error (MSE) loss function, which is formulated as:
\begin{equation}
\mathcal{L}_{\text{MSE}} = \frac{1}{N} \sum_{n=1}^{N} (\hat{s}_n - s_n)^2,
\end{equation}
where $\mathcal{L}_{\text{MSE}}$ denotes the MSE loss, $N$ denotes the total number of batch training samples, $s_n$ denotes the subjective score of the $i$-th ERP image.

\begin{table*}[]
\begin{threeparttable}
\caption{Performance comparison of the proposed VUGA and state-of-the-art OIQA models on OIQA databases. The best results are highlighted in bold.}
\renewcommand{\arraystretch}{1.5}
\setlength{\tabcolsep}{1.43mm}
\begin{tabular}{@{}cccccllllllllcccc@{}}
\toprule
\multirow{2}{*}{Method} & \multicolumn{2}{c}{CVIQ} & \multicolumn{2}{c}{OIQA} & \multicolumn{2}{l}{MVAQD} & \multicolumn{2}{l}{IQA-ODI} & \multicolumn{2}{c}{OSIQA} & \multicolumn{2}{l}{AIGCOIQA} & \multicolumn{2}{c}{JUFE-10K} & \multicolumn{2}{c}{OIQ-10K} \\ \cmidrule(l){2-17} 
                        & SRCC & PLCC & SRCC & PLCC & \multicolumn{1}{c}{SRCC} & \multicolumn{1}{c}{PLCC} & \multicolumn{1}{c}{SRCC} & \multicolumn{1}{c}{PLCC} & \multicolumn{1}{c}{SRCC} & \multicolumn{1}{c}{PLCC} & \multicolumn{1}{c}{SRCC} & \multicolumn{1}{c}{PLCC} & SRCC & PLCC & SRCC & PLCC \\ \cmidrule(r){1-17}
S-PSNR~\cite{yu2015framework} & 0.708 & 0.708 & 0.539 & 0.599 & - & - & - & - & - & - & - & - & 0.285 & 0.355 & 0.251 & 0.302 \\
WS-PSNR~\cite{sun2017weighted} & 0.610 & 0.672 & 0.526 & 0.581 & - & - & - & - & - & - & - & - & 0.284 & 0.353 & 0.248 & 0.295 \\
CPP-PSNR~\cite{zakharchenko2016quality} & 0.626 & 0.687 & 0.514 & 0.568 & - & - & - & - & - & - & - & - & 0.285 & 0.295 & 0.355 & 0.248 \\ 
WS-SSIM~\cite{chen2018spherical} & 0.911 & 0.929 & 0.503 & 0.504 & - & - & - & - & - & - & - & - & 0.248 & 0.388 & 0.062 & 0.223 \\ \cmidrule(r){1-17}
Assessor360~\cite{wu2024assessor360} & 0.964 & 0.977 & \textbf{0.980} & 0.975 & 0.961 & 0.972 & 0.957 & 0.963 & 0.533 & 0.832 & \textbf{0.914} & \textbf{0.912} & 0.690 & 0.694 & 0.773 & 0.790 \\
MC360IQA~\cite{sun2019mc360iqa} & 0.914 & 0.951 & 0.919 & 0.925 & 0.382 & 0.555 & 0.742 & 0.812 & 0.275 & 0.387 & 0.572 & 0.586 & 0.620 & 0.620 & 0.710 & 0.721 \\
VGCN~\cite{xu2020vgcn} & 0.965 & 0.963 & 0.958 & 0.951 & - & - & - & - & - & - & - & - & 0.464 & 0.367 & 0.698 & 0.755 \\
Fang22$\triangle$~\cite{fang2022perceptual} & 0.683 & 0.710 & 0.747 & 0.795 & 0.469 & 0.475 & 0.310 & 0.474 & 0.383 & 0.504 & 0.460 & 0.457 & 0.633 & 0.616 & 0.758 & 0.769 \\
MTAOIQA~\cite{yan2025matoiqa} & 0.962 & 0.965 & 0.960 & 0.972 & 0.592 & 0.642 & 0.335 & 0.528 & 0.646 & 0.792 & 0.456 & 0.432 & 0.821 & 0.822 & 0.824 & 0.829 \\
OIQAND~\cite{yan2024subjective} & 0.967 & 0.976 & 0.937 & 0.938 & 0.903 & 0.924 & 0.927 & 0.965 & 0.821 & 0.864 & 0.896 & 0.895 & 0.800 & 0.800 & 0.740 & 0.755 \\
Max360IQ~\cite{yan2025max360iq} & 0.966 & 0.970 & 0.922 & 0.919 & 0.724 & 0.769 & 0.797 & 0.731 & 0.862 & 0.898 & 0.409 & 0.505 & 0.563 & 0.491 & 0.751 & 0.733 \\
VU-BOIQA~\cite{yan2025vuboiqa} & 0.963 & 0.958 & 0.976 & 0.973 & 0.965 & 0.970 & 0.905 & 0.934 & 0.774 & 0.862 & 0.888 & 0.881 & 0.782 & 0.781 & 0.766 & 0.749 \\
VUGA & \textbf{0.970} & \textbf{0.984} & 0.973 & \textbf{0.974} & \textbf{0.966} & \textbf{0.973} & \textbf{0.967} & \textbf{0.978} & \textbf{0.928} & \textbf{0.950} & 0.913 & 0.868 & \textbf{0.846} & \textbf{0.842} & \textbf{0.830} & \textbf{0.834} \\ \bottomrule
\end{tabular}

\begin{tablenotes}
\item[]The $\triangle$ indicates a variant by removing the temporal hysteresis module.
\end{tablenotes}

\label{cmp:oiqa}
\end{threeparttable}
\end{table*}
\section{Experiments}
\label{sec:qu_ex}
\subsection{Experimental Settings. }

\textit{1) Databases:} We conduct extensive experiments on eight OIQA databases and five IQA databases. Among the selected OIQA databases, CVIQ~\cite{sun2018large}, OIQA~\cite{duan2018perceptual}, MVAQD~\cite{jiang2021cubemap} and IQA-ODI~\cite{yang2021spatial} are characterized by uniform distortion, while OSIQA~\cite{duan2023attentive} and JUFE-10K~\cite{yan2024subjective} contain non-uniform distortion. Furthermore, the OIQ-10K~\cite{yan2024omnidirectional} database has both non-uniformly and uniformly distorted OIs, and the AIGCOIQA database~\cite{yang2024aigcoiq2024} introduces AI-generated distortion. Regarding the test IQA databases, KADID-10k~\cite{lin2019kadid}, TID2013~\cite{tid}, and LIVE~\cite{live} are subject to synthetic distortion, whereas KonIQ-10k~\cite{hosu2020koniq} and SPAQ~\cite{fang2020cvpr} contain authentically distorted images.

\textit{2) Evaluation Metrics:} We adopt two metrics to evaluate the test models, including Pearson’s linear correlation coefficient (PLCC) and Spearman’s rank order correlation coefficient (SRCC). Higher values of these metrics indicate better performance. Following the suggestion in~\cite{vqeg}, a four-parameter logistic function is applied prior to calculating PLCC.

\textit{3) Implementation Details:} We randomly divide each database into a training set (80$\%$ images) and a test set (20$\%$ images). In the experiments described in Sections~\ref{sec:cmp_oiqa} and~\ref{sec:cmp_iqa}, the images are resized to 224$\times$224. In the other experiments, the images are resized to 1024$\times$1024. We freeze the parameters of the backbone, which has been pre-trained on ImageNet~\cite{deng2009imagenet}. We implement the proposed model using PyTorch on a high-performance server equipped with an Intel(R) Xeon(R) Gold 6326 CPU @ 2.90 GHz, a 24 GB NVIDIA GeForce RTX A5000 GPU, and 260 GB RAM. All experiments are trained for 10 epochs. The batch size is set to 8, and we use the Adam optimizer to optimize the proposed model. The learning rate is set to 1e-4 with a cosine annealing scheduler, and the weight decay is 1e-4.

\begin{table*}[]
\caption{Performance comparison of the proposed VUGA and state-of-the-art BIQA models on IQA databases. The best results are highlighted in bold.}
\renewcommand{\arraystretch}{1.5}
\centering
\begin{tabular}{@{}ccccccccccc@{}}
\toprule
                            & \multicolumn{2}{c}{KADID-10k}                               & \multicolumn{2}{c}{KonIQ-10k}                               & \multicolumn{2}{c}{LIVE}                                    & \multicolumn{2}{c}{SPAQ}                                    & \multicolumn{2}{c}{TID2013}                                     \\ \cmidrule(l){2-11} 
\multirow{-2}{*}{Method}    & SRCC                         & PLCC                         & SRCC                         & PLCC                         & SRCC                         & PLCC                         & SRCC                         & PLCC                         & SRCC                         & PLCC                         \\ \cmidrule(r){1-11}
DB-CNN~\cite{zhang2020blind}                    & 0.851                        & 0.856                        & 0.875                        & 0.884                        & 0.968                        & 0.971                        & 0.911                        & 0.915                        & 0.816                        & 0.865                        \\
HyperIQA~\cite{su2020blindly}                    & 0.852                        & 0.845                        & 0.906                        & 0.917                        & 0.962                        & 0.966                        & 0.911                        & 0.915                        & 0.840                        & 0.858                        \\
MUSIQ~\cite{ke2021musiq}                       & 0.875                        & 0.872                        & 0.916                        & 0.928                        & 0.940                        & 0.911                        & 0.918                        & 0.921                        & 0.773                        & 0.815                        \\
DEIQT~\cite{qin2023data}                       & 0.889                        & 0.887                        & 0.921                        & 0.934                        & 0.980                        & 0.982                        & 0.919                        & 0.923                        & 0.892                        & 0.908                        \\
LIQE~\cite{zhang2023blind}                        & 0.930                        & 0.931                        & 0.919                        & 0.908                        & 0.970                        & 0.951                        & -                            & -                            & -                            & -                            \\
LoDa~\cite{xu2023local}                        & 0.931                        & 0.936                        & 0.932                        & 0.944                        & 0.975                        & 0.979                        &\textbf{ 0.925} & \textbf{0.928}                        & 0.869                        & 0.901                        \\
TOPIQ~\cite{chen2024topiq}                        &  -                       &    -                     & 0.926                        & 0.939                        &  \textbf{0.984}                      &  0.984                      & 0.921 & 0.924                        &0.954                         & 0.958                        \\
VUGA & \textbf{0.973} &\textbf{0.975}  & \textbf{0.934} & \textbf{0.945} & 0.982 & \textbf{0.985} & 0.924 & 0.919 & \textbf{0.969} & \textbf{0.973} \\ \bottomrule
\end{tabular}
\label{cmp:iqa}
\end{table*}

\subsection{Experimental Results of BIQA models on the OIQA Databases}
\label{sec:cmp_oiqa}

Here, we evaluate the performance of four BIQA models, including HyperIQA~\cite{su2020blindly}, VCRNet~\cite{pan2022vcrnet} TOPIQ~\cite{chen2024topiq} and QualiClip~\cite{agnolucci2024quality}, where two input modes, \ie, ERP image and viewport sequence (marked by *) are set for better comparison. In addition, we test four baselines, including ResNet-50~\cite{he2016deep}, ResNet-152~\cite{he2016deep}, SwinV2-T~\cite{liu2022swin} and SwinV2-B~\cite{liu2022swin}, whose input is set as the ERP image.

As can be observed from Table~\ref{cross domain}, these general networks (including ResNet-50, ResNet-152, SwinV2-T and SwinV2-B) show competitive performance on these OIQA databases, which can be attributed to two reasons. The first is the \textit{easy} database issue~\cite{yan2025computational}. For these databases such as CVIQ, OIQA, MVAQD, IQA-ODI, and OSIQA, which are relatively small in scale and contain simple distortion types, allowing models like ResNet and Swin-Transformer to easily handle the distortions and achieve high performance with limited differences among models. However, they show relatively poor performance on these \textit{challenging} databases (\ie, AIGCOIQA, JUFE-10K and OIQ-10K). These results are consistent with the results of our previous study~\cite{yan2025computational}, where the difference lies in the input setting, \ie, the ERP image is used as input in this experiment while the extracted viewport sequence is used as input in our previous study~\cite{yan2025computational}. Second, although ERP images introduce geometric deformation, they still share a visual similarity to 2D planar images. This similarity enables models pre-trained on large-scale datasets to effectively transfer their learned hierarchical representations (\emph{e.g.}, edges, textures, and spatial structures) to the OIQA task~\cite{yan2025matoiqa}. Compared to their performance with viewport sequence as input, these general networks show slightly worse performance with raw ERP image as input; this is reasonable since there is no specific module being plugged in to deal with the inborn geometry deformation.

Among BIQA methods when using the raw ERP image or the viewport sequence as input, HyperIQA~\cite{su2020blindly} and VCRNet~\cite{pan2022vcrnet} exhibit consistently good performance. HyperIQA employs a self-adaptive hyper-network that dynamically generates image-specific parameters and integrates multi-scale and distortion-aware features to achieve content-adaptive quality prediction. VCRNet introduces a non-adversarial visual restoration framework, combined with optimized asymmetric residual blocks and an error-map–guided mixed loss, delivering robust assessment even under complex distortion. In contrast, TOPIQ~\cite{chen2024topiq} and QualiCLIP~\cite{agnolucci2024quality} perform well with the viewport sequence as input, but their performance drops markedly with the raw ERP image as input. This is because geometric deformation in the ERP image and non-uniform sampling disrupt TOPIQ’s top-down semantics-guided cross-scale attention and weaken QualiCLIP’s CLIP-based image–text alignment, which is learned primarily from 2D planar images. Although HyperIQA, VCRNet, and other BIQA models are promising, they show a similar tendency to the test general networks, indicating that these BIQA models can also be adapted to BOIQA with the sacrifice in performance. In summary, a natural gap exists between general networks or BIQA and BOIQA; however, it is not impossible to fill.

\subsection{Performance Comparison of OIQA Models}

Four FR-OIQA methods, including S-PSNR~\cite{yu2015framework}, WS-PSNR~\cite{sun2017weighted}, CPP-PSNR~\cite{zakharchenko2016quality} and WS-SSIM~\cite{chen2018spherical}, and Seven BOIQA methods, including Assessor360~\cite{wu2024assessor360}, MC360IQA~\cite{sun2019mc360iqa}, VGCN~\cite{xu2020vgcn}, Fang22~\cite{fang2022perceptual}, MTAOIQA~\cite{yan2025matoiqa}, OIQAND~\cite{yan2024subjective} and Max360IQ~\cite{yan2025max360iq}, are compared with the proposed VUGA on these OIQA databases. Note that these BOIQA models are retrained on eight OIQA databases with their default settings, respectively, and the input of VUBOIQA is a patch sequence with ten 224$\times$224 ERP patches and that of other models is a viewport sequence with eight 224$\times$224 viewports~\cite{yan2024subjective}.

\begin{table}[]
\caption{Ablation study results of each module in the VUGA model. w/o:without.}
\renewcommand{\arraystretch}{1.5}
\centering
\begin{tabular}{@{}ccccc@{}}
\toprule
\multirow{2}{*}{Method} & \multicolumn{2}{c}{\begin{tabular}[c]{@{}c@{}}JUFE-10K\end{tabular}} & \multicolumn{2}{c}{\begin{tabular}[c]{@{}c@{}}OIQ-10K\end{tabular}} \\ \cmidrule(l){2-5} 
                        & SRCC                                        & PLCC                                        & SRCC                                        & PLCC                                        \\ \cmidrule(r){1-5}
w/o CMP                & 0.835                                       & 0.832                                       & 0.819                                       & 0.822                                       \\
w/o SDA                 & 0.836                                       & 0.839                                       & 0.827                                       & 0.832                                       \\
w/o CAE                 & 0.842                                       & 0.839                                       & 0.828                                       & 0.833                                       \\
VUGA                    & \textbf{0.846}                                       & \textbf{0.842}                                       &  \textbf{0.830 }                                      & \textbf{0.834}                                       \\ \bottomrule
\end{tabular}
\label{ablation:component}
\end{table}

\begin{figure*}
\centering
\includegraphics[scale=0.60]{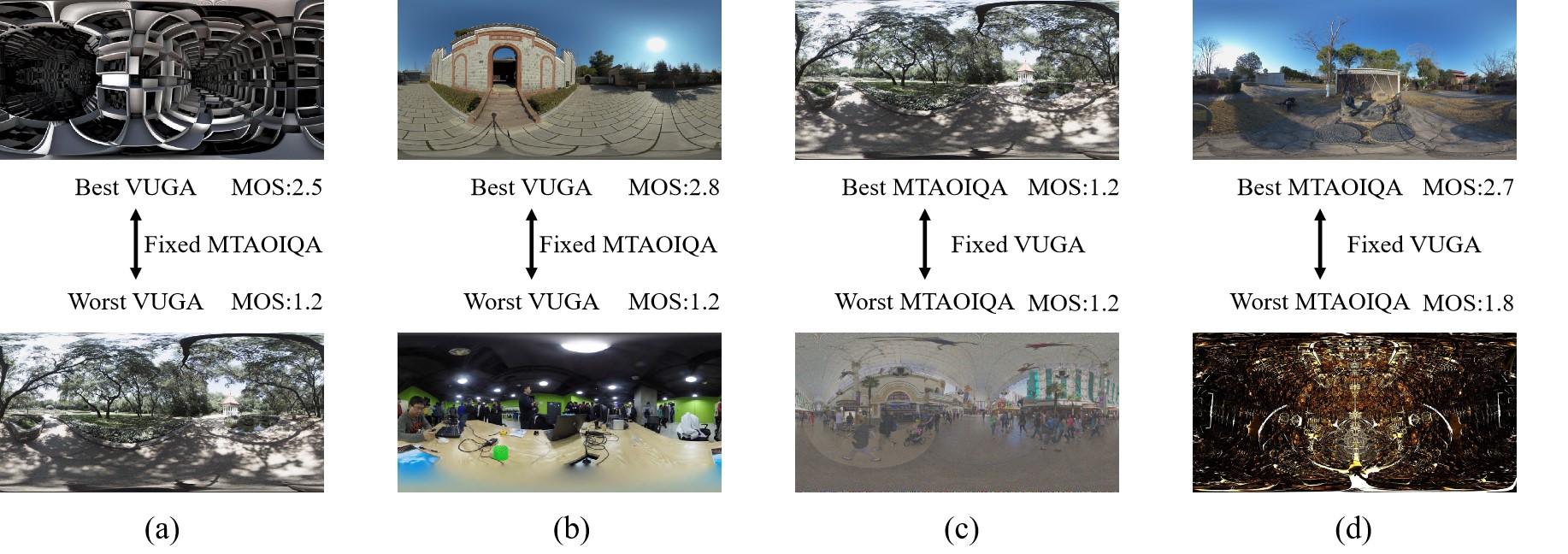}
\caption{gMAD competition results between MTAOIQA~\cite{yan2025matoiqa} and VUGA. The MOS of each image is shown in the bracket. (a) Fixed MTAOIQA at the low-quality level. (b) Fixed MTAOIQA at the high-guality level. (c) Fixed VUGA at the low-quality level. (d) Fixed VUGA at the high-quality level.}
\label{fig:gmad}
\end{figure*} 
\begin{figure*}
\centering
\includegraphics[scale=0.60]{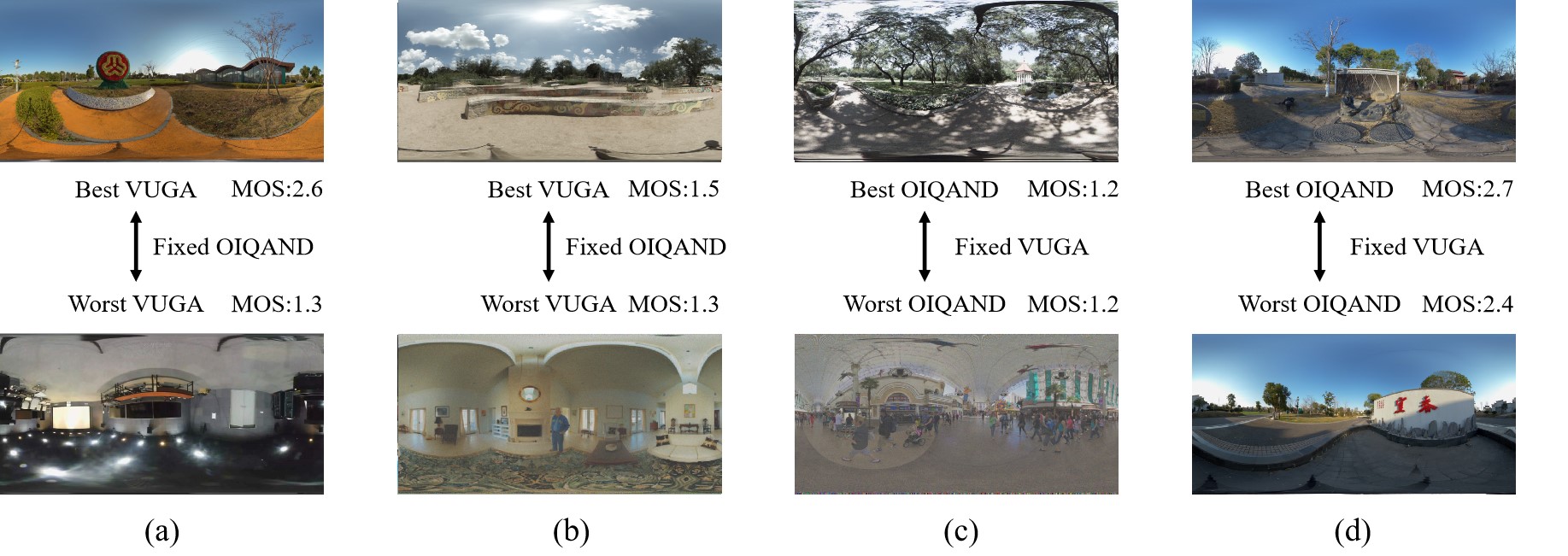}
\caption{gMAD competition results between OIQAND~\cite{yan2024subjective} and VUGA. The MOS of each image is shown in the bracket. (a) Fixed OIQAND at the low-quality level. (b) Fixed OIQAND at the high-quality level. (c) Fixed VUGA at the low-quality level. (d) Fixed VUGA at the high-auality level.}
\label{fig:gmad}
\end{figure*}

As shown in Table~\ref{cmp:oiqa}, FR-OIQA methods, \ie, S-PSNR, WS-PSNR, CPP-PSNR and WS-SSIM, consistently exhibit poor performance compared to these BOIQA models across all OIQA databases; this is primarily due to that FR-OIQA methods focus on pixel-level differences and overlook structural and semantic features, making them less effective for assessing the perceptual quality of OIs. For these deep learning-based BOIQA models, MTAOIQA outperforms other models on the JUFE-10K and OIQ-10K databases, whose main reason is that MTAOIQA optimizes itself by jointly training the main and auxiliary tasks, enabling it to more effectively capture non-uniform distortion. On other databases, Assessor360IQA outperforms other OIQA methods. This may be attributed to its unique recursive probability sampling (RPS) scheme, which enables it to capture more semantic information. Additionally, another possible reason is that its backbone (\ie, Swin Transformer) is better at perceiving distortion information compared to other networks. The proposed VUGA outperforms the second-best MTAOIQA in terms of SRCC and PLCC by 1.35$\%$ and 2.18$\%$, respectively, on the OIQ-10K and JUFE-10K databases. This advantage is attributed to the integration of DCN and local-global attention modeling, which effectively alleviates irregular geometric distortion caused by spherical projection and enhances the perception of local-global complex distortion. In particular, on the OSIQA database, VUGA improves SRCC and PLCC by 13.03$\%$ and 9.95$\%$, respectively, compared to Max360IQ. Overall, VUGA demonstrates exceptional performance across eight OIQA databases, with strong distortion perception capability and excellent generalizability.

\subsection{Experimental Results of the proposed VUGA on the IQA Databases}
\label{sec:cmp_iqa}

We further evaluate the adaptation ability of VUGA on five widely used IQA databases. Seven BIQA models are compared, including DB-CNN~\cite{zhang2020blind}, HyperIQA~\cite{su2020blindly}, MUSIQ~\cite{ke2021musiq}, DEIQT~\cite{qin2023data}, LIQE~\cite{zhang2023blind}, LoDa~\cite{xu2023local}, and TOPIQ~\cite{chen2024topiq}. The results of these methods are sourced from \cite{xu2023local} or their original papers. For the proposed VUGA, following ~\cite{fang2017no}, we repeat the training process 10 times with random data splits to mitigate performance bias and report the median SRCC and PLCC values. The comparison results are shown in Table~\ref{cmp:iqa}. From Table~\ref{cmp:iqa}, VUGA shows consistently superior performance on these IQA databases, outperforming all existing models in terms of SRCC and PLCC. In synthetic distortion databases such as KADID-10k and TID2013, VUGA significantly exceeds advanced distortion-aware methods, \ie, DEIQT, and LoDa, reflecting its enhanced capacity to model complex distortion patterns through effective global-local feature interaction. On the authentic distortion databases such as KonIQ-10k and SPAQ, VUGA also performs well, showing its strong ability to handle diverse scenes and realistic image distortion. Note that TOPIQ also achieves highly competitive results, which highlights the strength of spatially adaptive modeling. In contrast, VUGA maintains a top ranking with minimal variance, reflecting its better generalization capacity under cross-domain settings. We attribute this consistent advantage to its ERP-based representation, which preserves structural integrity while allowing localized quality inference. Unlike patch-based models constrained by viewport and patch sampling strategies, VUGA directly models the entire distortion field of OIs, thereby bridging spatial completeness and distortion sensitivity in a unified framework.

\subsection{Generalizability Validation}
\textit{1) Cross-Database Validation:} We conduct cross-database validation between JUFE-10K and OIQ-10K. In each setting, 80$\%$ of the images in one database are randomly selected for training, and the other database is used entirely for testing. The experimental results are summarized in Table~\ref{cross database}. From Table~\ref{cross database}, we can clearly observe that VUGA outperforms all compared methods, with improvements of SRCC and PLCC exceeding 12$\%$ and 10$\%$, respectively, compared to the second-best model, \ie, OIQAND. When trained on OIQ-10K, VUGA achieves notably better performance, which can be attributed to the richer distortion diversity in this database, as it contains both uniform and non-uniform distortions. Overall, these findings confirm the strong generalizability of VUGA across different distortion distributions.

\textit{2) The gMAD Competition:} To further compare the generalizability of these test models, the group Maximum Differentiation (gMAD) competition methodology~\cite{ma2018group}~\cite{wang2008maximum}~\cite{yan2021exposing} is employed as a diagnostic framework. This adversarial evaluation paradigm aims to identify latent deficiencies through a dynamic competitive mechanism. Within this framework, participating models are configured to perform dual adversarial roles: 1) Attacker, where conflicting samples are systematically extracted from opponents' predictions based on score-level similarity metrics while maintaining significant perceptual discrepancies, thus revealing evaluation blind spots; and 2) Defender, where the model demonstrates assessment consistency by resisting adversarial challenges. Through comparative analysis of these discriminative instances, fundamental divergences in visual quality perception are demonstrated.

Here, we compare VUGA with OIQAND~\cite{yan2024subjective} and MTAOIQA~\cite{yan2025matoiqa}. All models are trained on the JUFE-10K database and evaluated on the OIQ-10K database. As shown in Fig.~\ref{fig:gmad}, when VUGA acts as the defender, the image pairs selected by the attacking models exhibited an average subjective quality difference of not more than 0.4. In contrast, when VUGA acts as an attacker, it successfully identifies image pairs with subjective quality differences greater than 1.2, while the competing models give similar quality scores. These results demonstrate the superior discriminative power and generalizability of VUGA.

\begin{table}[h]
\caption{Cross database validation. The best results are highlighted in bold.}
\renewcommand{\arraystretch}{1.5}
\centering
\begin{tabular}{@{}ccccc@{}}
\toprule
\multirow{2}{*}{Method} & \multicolumn{2}{c}{\begin{tabular}[c]{@{}c@{}}Train JUFE-10K\\ Test OIQ-10K\end{tabular}} & \multicolumn{2}{c}{\begin{tabular}[c]{@{}c@{}}Train OIQ-10K\\ Test JUFE-10K\end{tabular}} \\ \cmidrule(l){2-5} 
                        & SRCC                                        & PLCC                                        & SRCC                                        & PLCC                                        \\ \cmidrule(r){1-5}
MC360IQA~\cite{sun2019mc360iqa}                & 0.278                                       & 0.290                                       & 0.253                                       & 0.319                                       \\
VGCN~\cite{xu2020vgcn}                    & 0.418                                       & 0.426                                       & 0.517                                       & 0.550                                       \\
Fang22~\cite{fang2022perceptual}                  & 0.162                                       & 0.274                                       & 0.366                                       & 0.429                                       \\
Assessor360~\cite{wu2024assessor360}             & 0.367                                       & 0.357                                       & 0.614                                       & 0.624                                       \\
VU-BOIQA~\cite{yan2025vuboiqa}                & 0.472                                       & 0.458                                       & 0.610                                       & 0.625                                       \\
MTAOIQA~\cite{yan2025matoiqa}             & 0.469                                       & 0.468                                       & 0.696                                       &  0.708                                       \\
OIQAND~\cite{yan2024subjective}             & 0.536                                       & 0.529                                       & 0.301                                       &  0.337                                       \\
VUGA                    &\textbf{0.603}                                       & \textbf{0.594}                                       &\textbf{0.713}                                       &\textbf{ 0.718}                                       \\ \bottomrule
\end{tabular}
\label{cross database}
\end{table}

\subsection{Ablation Studies}
In this subsection, we perform two ablation experiments on the JUFE-10K and OIQ-10K databases.

\textit{1) Effectiveness of Individual Module:} We test the performance of VUGA by separately removing the CMP, SDA, and CAE modules. As shown in Table~\ref{ablation:component}, we can observe that removing the CMP module results in the most significant degradation. This shows that the CMP module, through the combination of DCN and a local-global attention mechanism, effectively alleviates irregular geometric distortion in the ERP image and improves the model’s ability to perceive local-global complex distortion. The absence of the SDA module causes consistent performance drops on both databases, highlighting its role in multiscale feature alignment and distortion-aware fusion. Furthermore, although the removal of the CAE module leads to a relatively smaller decline, it still affects the network’s capacity to refine perceptual features during the regression stage. Overall, each module plays a positive role in the overall architecture.

\begin{table}[]
\caption{Performance by varying different input resolutions.}
\renewcommand{\arraystretch}{1.5}
\centering
\begin{tabular}{@{}ccccc@{}}
\toprule
\multirow{2}{*}{Resolution} & \multicolumn{2}{c}{JUFE-10K} & \multicolumn{2}{c}{OIQ-10K} \\ \cmidrule(l){2-5} 
                            & SRCC          & PLCC         & SRCC         & PLCC         \\ \cmidrule(r){1-5}
224$\times$224                        &  0.810             & 0.808            &0.783              &  0.793            \\
512$\times$512                         &  0.828             & 0.825             & 0.817             &  0.821            \\
768$\times$768                         &  0.835             &  0.832             &  0.829           & 0.834             \\
1024$\times$1024                        &\textbf{0.846}               &\textbf{0.842}              &\textbf{0.830}              & \textbf{0.834}             \\ \bottomrule
\end{tabular}
\label{ablation:size}
\end{table}

\textit{2) Impact of Image Resolution:} We evaluate the performance of VUGA by varying different input resolutions on the JUFE-10K and OIQ-10K databases. As shown in Table~\ref{ablation:size}, VUGA performs best when the input resolution is set to 1024$\times$1024. As resolution decreases, the overall performance of VUGA also decreases. This trend is reasonable, since if the input image is too small, it can cause the model to struggle with capturing distortion.

\section{Conclusion}
\label{sec:conc}

In this paper, we present a novel solution to narrow the natural gap between BIQA and BOIQA, which has been less investigated in the IQA research community. Specifically, we first find that those backbones for general computer vision tasks and BIQA models can be easily adapted to BOIQA, in which the OIs with inborn unnatural geometry deformation can be treated as 2D planar images, and their performance is also relatively acceptable. Furthermore, we design a new BOIQA model, namely VUGA, which is \textit{viewport-unaware}, \textit{unified}, and \textit{generalized}, \ie, it accepts the raw ERP image as input, can be applied to BIQA and shows good generalizability. To achieve this, we elaborately design a CMP module for distilling quality-aware features from the inborn geometric structures and an AFF module for merging those hierarchical quality-sensitive features in a progressive manner.

%

\ifCLASSOPTIONcaptionsoff
  \newpage
\fi



%

\bibliographystyle{IEEEtran}
\bibliography{egbib}





\begin{IEEEbiography}[{\includegraphics[width=1.0in,height=1.3in,clip]{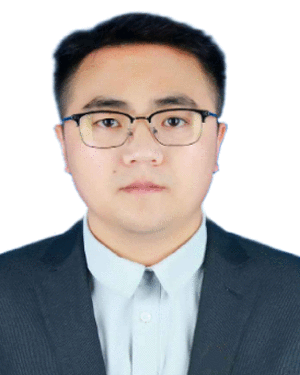}}]{Jiebin Yan} received the Ph.D. degree from the Jiangxi University of Finance and Economics, Nanchang, China. He was a Computer Vision Engineer with MTlab, Meitu. Inc, and Research Intern with MOKU Laboratory, Alibaba Group. From 2021 to 2022, he was a visiting Ph.D. student with the Department of Electrical and Computer Engineering, University of Waterloo, Canada. He is currently a Lecturer with the School of Computing and Artificial Intelligence, Jiangxi University of Finance and Economics. His research interests include visual quality assessment and computer vision.
\end{IEEEbiography}

\begin{IEEEbiography}[{\includegraphics[width=1.0in,height=1.3in,clip]{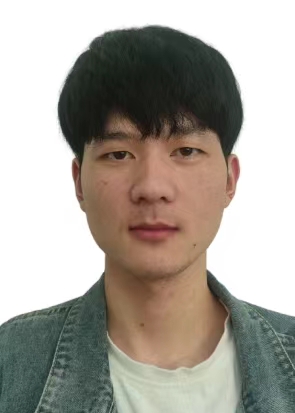}}]{Kangcheng Wu} received the B.E. degree from the Jiangxi University of Finance and Economics, Nanchang, China, in 2023. He is currently working toward the M.S. degree with the School of Computing and Artificial Intelligence, Jiangxi University of Finance and Economics, Nanchang. His research interests include visual quality assessment and VR image processing.
\end{IEEEbiography}

\begin{IEEEbiography}[{\includegraphics[width=1.0in,height=1.3in,clip]{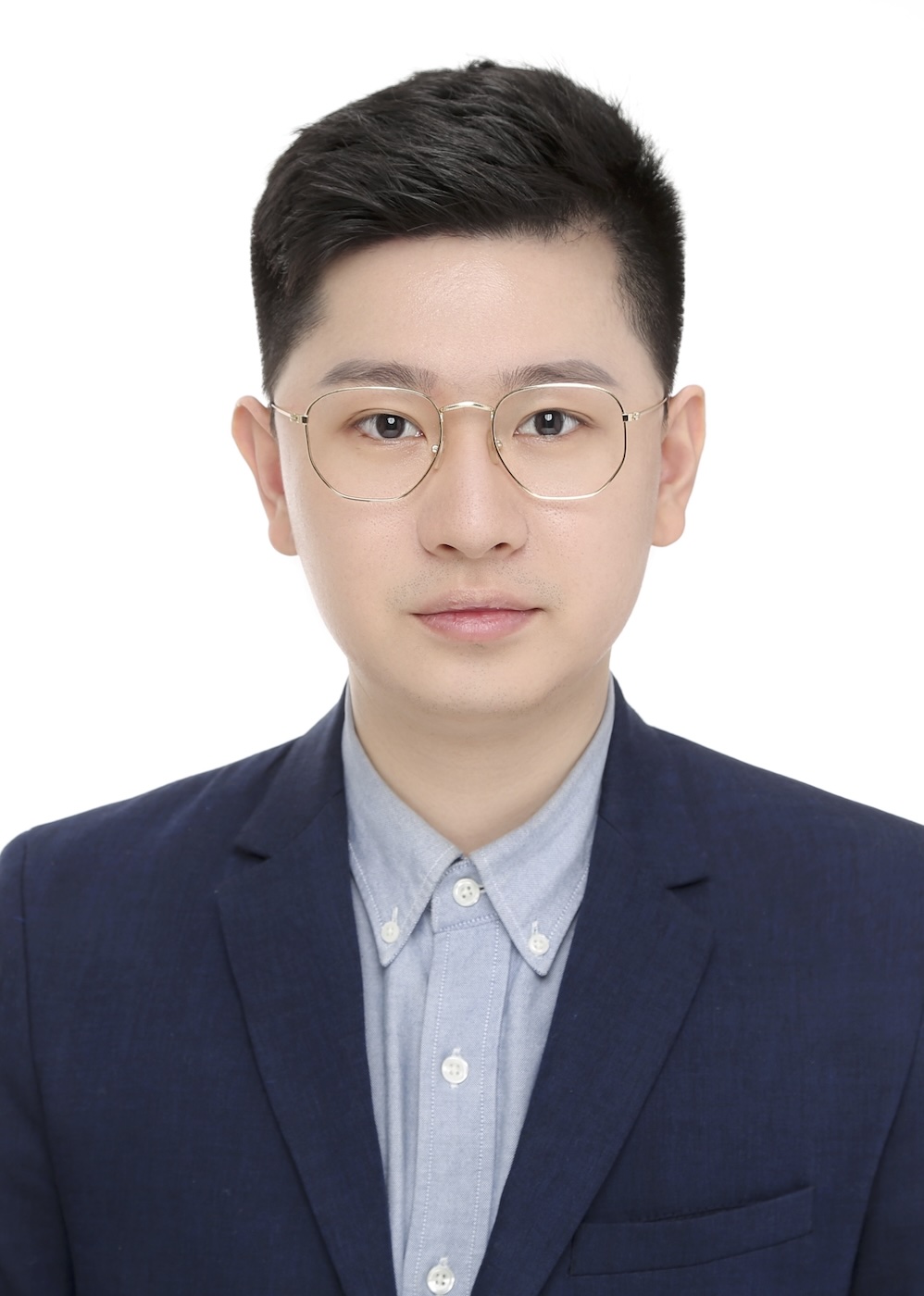}}]{Jingwen Hou} received the bachelor’s degree from Beijing University of Posts and Telecommunications, Beijing, China, the master’s degree from Carnegie Mellon University, Pittsburgh, USA, both in E-business, and the Ph.D. degree from Nanyang Technological University, Singapore. He is currently with the School of Computing and Artificial Intelligence, Jiangxi University of Finance and Economics, Nanchang, China. His research interests include multimedia quality assessment and computer vision.
\end{IEEEbiography}

\begin{IEEEbiography}[{\includegraphics[width=1.0in,height=1.3in,clip]{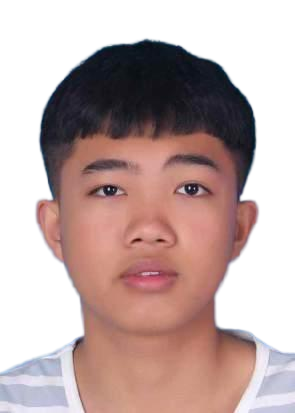}}]{Jiayu Zhang} received the B.E. degree from Xi'an University of Posts $\&$ Telecommunications, Xi'an, China, in 2023. He is currently working toward the M.S. degree with the School of Computing and Artificial Intelligence, Jiangxi University of Finance and Economics, Nanchang. His research interests include visual quality assessment and VR image processing.
\end{IEEEbiography}

\begin{IEEEbiography}[{\includegraphics[width=1.0in,height=1.3in,clip]{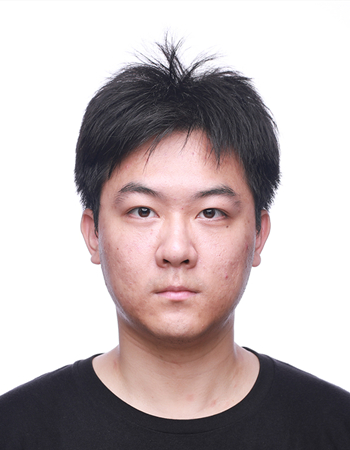}}]{Pengfei Chen} received the BS degree from Xidian University, Xi’an, China, in 2014, and the PhD degree from the China University of Mining and Technology, Xuzhou, China, in 2022. He is currently a lecturer with the School of Artificial Intelligence, Xidian University. His research interests include image/video quality assessment, video quality of experience, action quality assessment and domain adaptation/generalization.
\end{IEEEbiography}

\vspace{-130 mm}

\begin{IEEEbiography}[{\includegraphics[width=1.0in,height=1.3in,clip]{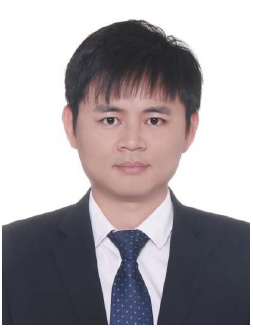}}]{Yuming Fang}(S’13–SM’17) received the B.E. degree from Sichuan University, Chengdu, China, the M.S. degree from the Beijing University of Technology, Beijing, China, and the Ph.D. degree from Nanyang Technological University, Singapore. He is currently a Professor with the School of Computing and Artificial Intelligence, Jiangxi University of Finance and Economics, Nanchang, China. His research interests include visual attention modeling, visual quality assessment, computer vision, and 3D image/video processing. He serves on the editorial board for \textsc{IEEE Transactions on Multimedia} and \textsc{Signal Processing: Image Communication}.
\end{IEEEbiography}

\end{document}